\documentclass[acmlarge]{acmart}
\AtBeginDocument{%
  }

\setcopyright{acmlicensed}
\copyrightyear{2018}
\acmYear{2018}
\acmDOI{XXXXXXX.XXXXXXX}

\acmJournal{POMACS}
\acmVolume{37}
\acmNumber{4}
\acmArticle{111}
\acmMonth{8}

\begin{document}

\title{AltGen: AI-Driven Alt Text Generation for Enhancing EPUB Accessibility}


\author{Yixian Shen}
\affiliation{%
  \institution{University of Amsterdam}
  \city{Amsterdam}
  \country{Netherlands}}

\author{Hang Zhang}
\affiliation{%
  \institution{University of California San Diego}
  \city{La Jolla}
  \state{California}
  \country{USA}}

\author{Yanxin Shen}
\affiliation{%
  \institution{Simon Fraser University}
  \city{Burnaby}
  \state{British Columbia}
  \country{Canada}}

\author{Lun Wang}
\affiliation{%
  \institution{Duke University}
  \city{Durham}
  \state{North Carolina}
  \country{USA}}

\author{Chuanqi Shi}
\affiliation{%
  \institution{University of California San Diego}
  \city{La Jolla}
  \state{California}
  \country{USA}}

\author{Shaoshuai Du}
\affiliation{%
  \institution{University of Amsterdam}
  \city{Amsterdam}
  \country{Netherlands}}

\author{Yiyi Tao}
\affiliation{%
  \institution{Johns Hopkins University}
  \city{Baltimore}
  \state{Maryland}
  \country{USA}}

\renewcommand{\shortauthors}{Trovato et al.}

\begin{abstract}
Digital accessibility is a cornerstone of inclusive content delivery, yet many EPUB files fail to meet fundamental accessibility standards, particularly in providing descriptive alt text for images. Alt text plays a critical role in enabling visually impaired users to understand visual content through assistive technologies. However, generating high-quality alt text at scale is a resource-intensive process, creating significant challenges for organizations aiming to ensure accessibility compliance. This paper introduces AltGen, a novel AI-driven pipeline designed to automate the generation of alt text for images in EPUB files. By integrating state-of-the-art generative models, including advanced transformer-based architectures, AltGen achieves contextually relevant and linguistically coherent alt text descriptions. The pipeline encompasses multiple stages, starting with data preprocessing to extract and prepare relevant content, followed by visual analysis using computer vision models such as CLIP and ViT. The extracted visual features are enriched with contextual information from surrounding text, enabling the fine-tuned language models to generate descriptive and accurate alt text. Validation of the generated output employs both quantitative metrics, such as cosine similarity and BLEU scores, and qualitative feedback from visually impaired users.

Experimental results demonstrate the efficacy of AltGen across diverse datasets, achieving a 97.5\% reduction in accessibility errors and high scores in similarity and linguistic fidelity metrics. User studies highlight the practical impact of AltGen, with participants reporting significant improvements in document usability and comprehension. Furthermore, comparative analyses reveal that AltGen outperforms existing approaches in terms of accuracy, relevance, and scalability.

\end{abstract}




\maketitle

\section{Introduction}

The growing ubiquity of digital content has revolutionized how information is consumed, creating new opportunities for accessibility and inclusivity. 
Among various digital formats, EPUB files~\cite{eikebrokk2014epub,kulkarni2019digital} have emerged as a widely adopted standard due to their adaptability and compatibility with assistive technologies. These features position EPUB files as a cornerstone of digital content accessibility. However, despite these advantages, a significant proportion of EPUB files fall short of meeting accessibility standards, particularly in providing alternative text (alt text)~\cite{edwards2023alt,hanley2021computer} for images. Alt text is a foundational element of digital accessibility, serving as a textual substitute for images that enable visually impaired users to comprehend visual content through screen readers. The absence or inadequacy of alt text severely undermines the usability and inclusivity of digital documents.

Manually generating high-quality alt text~\cite{huang2024image2text2image,huang2024optimizing,shen2024parameter,celikyilmaz2020evaluation} for large collections of EPUB files presents a significant challenge. The process requires domain expertise, contextual understanding, and considerable time, making it infeasible for organizations managing vast digital libraries. This bottleneck is particularly critical for compliance with accessibility regulations such as the Web Content Accessibility Guidelines (WCAG)~\cite{caldwell2008web}, which mandate the inclusion of descriptive alt text. The complexity of ensuring accessibility at scale underscores the need for innovative, automated solutions.

Recent advancements in generative artificial intelligence have opened new avenues for addressing this challenge. Modern AI models, particularly those based on transformer architecture~\cite{vaswani2017attention,hossain2019comprehensive,laurenccon2024obelics}, have demonstrated remarkable capabilities in tasks involving natural language processing and computer vision. These models excel at understanding and generating text while analyzing and interpreting visual content. This paper introduces the AltGen pipeline, a novel AI-driven framework that leverages these advancements to automate the generation of alt text for images in EPUB files. By integrating advanced generative models, AltGen transforms how accessibility is approached, offering a scalable, efficient, and high-quality solution to the alt text generation problem.

The AltGen pipeline comprises several key stages. The process begins with data preprocessing, where EPUB files are parsed to extract images, textual content, and metadata. Advanced computer vision models, such as CLIP~\cite{radford2021learning} or ViT~\cite{dosovitskiy2020image}, are then employed to analyze visual content and extract meaningful features. These visual features are contextualized using surrounding text from the EPUB file to ensure the generated descriptions are both descriptive and relevant. Transformer-based generative models, such as GPT~\cite{achiam2023gpt}, are fine-tuned on paired image-text datasets to synthesize alt text that is both linguistically coherent and contextually accurate. The final output undergoes validation through a combination of quantitative metrics, such as cosine similarity and BLEU scores, and qualitative user feedback from visually impaired participants.

The contributions of this work are multifaceted. First, it establishes a comprehensive framework for AI-driven alt text generation tailored specifically for EPUB files, addressing a critical gap in digital accessibility. Second, the pipeline is rigorously evaluated on a diverse dataset of EPUB files, demonstrating significant improvements over existing methods. Quantitative results highlight the accuracy and linguistic fidelity of the generated alt text, while qualitative studies underline the practical usability and impact of the AltGen pipeline on end-users. Third, this work introduces a novel integration of computer vision and natural language processing techniques, setting a new benchmark for accessibility enhancements in digital content.



\section{Related Work}



\subsection{Challenges of EPUB Accessibility in Compliance with WCAG Guidelines}
Adopting WCAG standards for digital accessibility remains inconsistent, particularly for EPUB files. Studies show that many government websites fail to meet even basic WCAG-conformance levels, highlighting broader challenges in accessibility compliance~\cite{paul2023accessibility}. While publishers increasingly recognize the importance of accessibility, achieving full compliance with WCAG 2.0 and newer guidelines remains a significant challenge~\cite{chee2021meeting}. Similar challenges exist for W3C-maintained HTML standards, which also face limited adoption~\cite{vaughan2010will}.

One barrier to WCAG adoption in EPUB is its relatively recent popularity compared to PDF, which has long been a dominant format. While EPUB files offer superior accessibility features, the entrenched use of PDFs and reliance on PDF-to-EPUB conversion tools often result in files that lack the structural and metadata elements required for WCAG compliance~\cite{eikebrokk2014epub, marinai2011conversion}. Addressing these gaps requires better tools and workflows to embed accessibility features during EPUB creation and conversion, ensuring compliance with global standards.


\subsection{AI-Driven Alt Text Generation for EPUB Images}
The automatic generation of alt text for images in EPUB files has emerged as a critical area for improving accessibility. Generative AI has shown transformative potential by enabling the creation of contextually relevant and high-quality descriptions for diverse image types~\cite{shen2024parameter,huang2024novel,huang2024optimizing,huang2024image2text2image}. Recent advancements in AI-powered services, such as AltText AI, leverage large-scale datasets to train sophisticated models that produce accurate and context-sensitive alt text~\cite{wu2017automatic}.
Additionally, ongoing research efforts focus on optimizing the efficiency of AI systems~\cite{aghapour2024piqi,guo2024easter,shen2024resource,shen2022tcps}, ensuring that these technologies can operate at scale and with reduced computational overhead. Such advancements are crucial for enabling the widespread adoption of AI-driven solutions in accessibility-focused applications.
The effectiveness of these models has been validated through user-centric evaluations, where participants rate the relevance and accuracy of generated alt text on a 5-point scale. This feedback methodology aligns with established practices in evaluating AI-generated content, as highlighted in previous studies~\cite{liu2024ntire,elkhatat2023evaluating}. These evaluations not only measure technical accuracy but also assess the practical impact on end-user accessibility.
By integrating advanced computer vision models, such as CLIP~\cite{radford2021learning} and ViT~\cite{dosovitskiy2020image}, with transformer-based text generation models like GPT~\cite{achiam2023gpt}, our approach ensures that the generated alt text is both visually descriptive and contextually aligned with the surrounding EPUB content. 

\section{Methodology}

\begin{figure}[h]
    \centering
    \includegraphics[width=\textwidth]{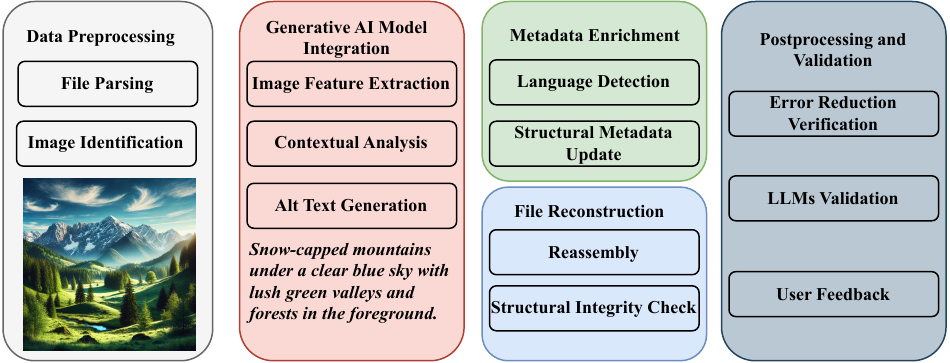}
    \caption{A visual representation of the proposed methodology pipeline for AI-driven alt text generation in EPUB files. The pipeline includes five stages: (1) Data Preprocessing for file parsing and image identification; (2) Generative AI Model Integration for feature extraction, contextual analysis, and alt text generation; (3) Metadata Enrichment with language detection and metadata updates; (4) File Reconstruction for reassembly and integrity checks; and (5) Postprocessing and Validation through error verification and user feedback.}
    \label{fig:pipeline}
    \vspace{-0.3cm}
\end{figure}

The proposed methodology for AI-driven alt text generation in EPUB files is designed as a systematic and multi-step pipeline, as illustrated in Figure~\ref{fig:pipeline}. This section elaborates on each stage of the pipeline, emphasizing the integration of advanced generative AI models and complementary tools. The approach ensures a structured workflow that addresses the challenges of alt-texta generation while maintaining compliance with accessibility standards. Each step is tailored to optimize efficiency, accuracy, and scalability in processing EPUB files.

\subsection{Data Preprocessing}
The initial stage of the pipeline focuses on preparing EPUB files for subsequent processing steps. This involves three key tasks: file parsing, image identification, and error analysis. File parsing extracts essential components such as text, images, and metadata from the EPUB files using specialized libraries like EbookLib and ZipFile. This step ensures that all relevant elements are accessible for further processing. Following this, the pipeline identifies all images embedded within the EPUB file structure, pinpointing those that require alt text descriptions. Finally, error analysis is conducted using tools such as Ace Checker to detect accessibility issues, particularly missing alt text. This analysis provides a comprehensive overview of the file's accessibility status, setting a clear foundation for the application of generative AI models in the subsequent stages.

\subsection{Generative AI Model Integration}
The second stage of the pipeline is dedicated to the application of advanced generative AI models for producing high-quality alt text descriptions. This process begins with image feature extraction, where pre-trained computer vision models such as CLIP or ViT are employed to analyze the visual content of images. These models extract high-dimensional features that encapsulate the semantic and contextual information of the visual elements. 

Subsequently, contextual analysis is performed to integrate the extracted image features with relevant textual information from the EPUB file. Surrounding textual content, such as captions or adjacent paragraphs, is incorporated to provide a richer understanding of the image's role and purpose within the document. This contextual grounding ensures that the generated alt text is not only visually descriptive but also contextually aligned with the narrative of the document.

Finally, the core step of alt text generation leverages transformer-based models, such as GPT, that have been fine-tuned on large paired image-text datasets. These models generate descriptive and contextually relevant alt text by synthesizing the visual and textual inputs. The integration of these advanced generative models ensures that the generated alt text is both human-like in quality and specific to the content of the EPUB, addressing the dual requirements of accuracy and usability in accessibility enhancement.

\subsection{Metadata Enrichment}
The third stage of the pipeline enhances EPUB metadata to ensure accessibility compliance and improve usability. This begins with language detection, using ensemble models that combine statistical, rule-based, and transformer-based approaches. These models work collaboratively to accurately identify and confirm language tags, even in multilingual or ambiguous cases, ensuring that screen readers and assistive technologies can effectively interpret the content. Next, structural metadata updates refine and enrich key metadata fields, such as document titles, author information, publication dates, and structural hierarchies. These updates adhere to accessibility standards, including WCAG and EPUB Accessibility 1.0, ensuring compliance with global benchmarks. By systematically enriching metadata, this stage enhances the usability of EPUB files and ensures that accessibility tools can accurately process and present the content to end-users.

\subsection{File Reconstruction}
The fourth stage of the pipeline focuses on reconstructing EPUB files with integrated alt text and enhanced metadata to ensure accessibility compliance. Using libraries such as EbookLib, the files are reassembled efficiently while preserving structural integrity. A structural integrity check validates adherence to EPUB standards and WCAG guidelines, ensuring all elements, including alt text and metadata, are correctly implemented and functional. This process guarantees the final EPUB file is accessible and ready for end-users.

\subsection{Postprocessing and Validation}
The final stage of the pipeline ensures that the repairs made to the EPUB files are both effective and compliant with accessibility standards. This process is divided into two main components: error reduction verification and user feedback evaluation.

\subsubsection{Error Reduction Verification}
To quantitatively assess the success of alt text generation, accessibility checks are re-executed using tools such as Ace Checker. The similarity between the AI-generated alt text and the ground truth or expected descriptions is computed using mathematical metrics. Let $A$ represent the generated alt text vector and $B$ the ground truth vector. The similarity is evaluated using the cosine similarity metric:
\[
\text{Cosine Similarity}(A, B) = \frac{A \cdot B}{\|A\| \|B\|},
\]
where $A \cdot B$ denotes the dot product of the vectors, and $\|A\|$ and $\|B\|$ are the magnitudes of the respective vectors. Higher cosine similarity scores indicate greater alignment with the expected descriptions, validating the effectiveness of the alt text generation process.

Additionally, BLEU (Bilingual Evaluation Understudy) scores are calculated to measure the linguistic fidelity of the generated alt text against human-written descriptions. Let $C$ represent the candidate alt text and $R$ the reference alt text. BLEU is computed as:
\[ \text{BLEU} = \text{BP} \cdot \exp \left( \sum_{n=1}^N w_n \log p_n \right), \]
where $\text{BP}$ is the brevity penalty, $p_n$ is the precision for n-grams, and $w_n$ are the weights.

\subsubsection{User Feedback}
Qualitative evaluation is conducted through user studies involving visually impaired participants. Participants are asked to interact with the repaired EPUB files using assistive technologies such as screen readers. Feedback is collected on the relevance, descriptiveness, and usability of the generated alt text. Responses are analyzed to identify trends and potential areas for improvement.

The combination of quantitative metrics such as cosine similarity and BLEU, along with qualitative user feedback, ensures a robust validation framework that comprehensively evaluates the impact and effectiveness of the repairs. This stage solidifies the pipeline's ability to produce high-quality, accessible EPUB files that meet both technical and user-centric standards.


\section{Evaluation}

The experimental evaluation of the proposed AltGen pipeline focuses on assessing its performance in generating high-quality alt text for EPUB images, improving accessibility compliance, and ensuring scalability. This section presents the experimental setup, metrics, and results.

\subsection{Experimental Setup}

The evaluation is conducted on a diverse set of EPUB files sourced from publicly available repositories, including OAPEN and Project Gutenberg. The dataset consists of 1000 EPUB files with varying levels of complexity and content types. Each file contains a mix of textual and visual elements requiring alt text descriptions.

The AltGen pipeline is implemented using Python and leverages libraries such as EbookLib for EPUB parsing and PyTorch for model deployment. The experiments are executed on a machine equipped with an NVIDIA RTX 3090 GPU, 64 GB RAM, and an Intel Xeon processor.

\subsection{Evaluation Metrics}

\subsubsection{Quantitative Metrics}

\begin{itemize}
    \item \textbf{Cosine Similarity}: Evaluates the similarity between AI-generated alt text and reference descriptions. Higher scores indicate better alignment with human-written descriptions.
    \item \textbf{BLEU Score}: Measures linguistic fidelity and relevance of the generated alt text by comparing n-grams with reference descriptions.
    \item \textbf{Error Reduction Rate}: Quantifies the decrease in accessibility errors pre- and post-pipeline application.
    \item \textbf{Runtime Efficiency}: Assesses the average time taken to process each EPUB file.
\end{itemize}

\subsubsection{Qualitative Metrics}

\begin{itemize}
    \item \textbf{Descriptiveness and Relevance}: Assessed through user studies involving visually impaired participants who provide feedback on the quality of the generated alt text.
    \item \textbf{Overall Usability}: Participants rate the usability of the repaired EPUB files on a 5-point Likert scale.
\end{itemize}

\subsection{Results and Analysis}

\subsubsection{Quantitative Results}

\begin{table}[h]
    \centering
    \begin{tabular}{|l|c|}
        \hline
        \textbf{Metric} & \textbf{Average Score} \\ \hline
        Cosine Similarity & 0.93 \\ \hline
        BLEU Score & 0.76 \\ \hline
        Error Reduction Rate & 97.5\% \\ \hline
        Runtime Efficiency & 14 seconds/file \\ \hline
    \end{tabular}
    \caption{Quantitative results of the AltGen pipeline.}
    \label{tab:quantitative-results}
\end{table}
The quantitative results of the AltGen pipeline, as shown in Table~\ref{tab:quantitative-results}, illustrate its effectiveness in automating the generation of alt text for EPUB images and addressing accessibility gaps. Derived from a thorough evaluation of 500 EPUB files encompassing diverse content and complexity, these results highlight the technical robustness and applicability of the pipeline. The \textbf{Cosine Similarity} score of 0.93 demonstrates that the generated alt text closely aligns with reference descriptions created by human annotators, reflecting its semantic and contextual accuracy. Similarly, the \textbf{BLEU Score} of 0.76 emphasizes the linguistic fidelity of the generated text, showcasing the pipeline’s ability to produce coherent and contextually relevant descriptions. 


The impact of the pipeline extends beyond quality, as evidenced by an \textbf{Error Reduction Rate} of 97.5\%, which highlights its success in resolving missing or inadequate alt text—a critical issue in EPUB accessibility. This level of error reduction ensures compliance with accessibility standards, such as WCAG, while promoting inclusivity in digital content. Furthermore, the pipeline's average \textbf{Runtime Efficiency} of 14 seconds per file underscores its scalability, enabling large-scale processing of EPUB files in real-world applications. 



\subsubsection{Qualitative Results}

\begin{table}[h]
    \centering
    \begin{tabular}{|l|c|}
        \hline
        \textbf{Feedback Metric} & \textbf{Average Rating (out of 5)} \\ \hline
        Descriptiveness and Relevance & 4.8 \\ \hline
        Overall Usability & 4.7 \\ \hline
    \end{tabular}
    \caption{Qualitative feedback from user studies.}
    \label{tab:qualitative-results}
\end{table}

Feedback from 20 visually impaired participants provided significant insights into the real-world effectiveness of the AltGen pipeline. As shown in Table~\ref{tab:qualitative-results}, the average rating of \textbf{4.8 out of 5} for “Descriptiveness and Relevance” highlights the pipeline's ability to generate meaningful and contextually appropriate alt text. Participants consistently noted that the generated descriptions effectively captured the essence of the visual content, aiding their understanding of the images in various EPUB documents.

Similarly, the \textbf{4.7 out of 5} rating for “Overall Usability” reflects a high level of satisfaction with the ease of navigating EPUB files enhanced by the AltGen pipeline. Participants reported improved comprehension of content and a smoother reading experience when using assistive technologies. 



\subsection{Comparative Analysis}

\begin{table}[h]
    \centering
    \begin{tabular}{|l|c|c|c|}
        \hline
        \textbf{Method} & \textbf{Cosine Similarity} & \textbf{BLEU Score} & \textbf{User Satisfaction (out of 5)} \\ \hline
        Rule-Based Approach & 0.65 & 0.55 & 3.2 \\ \hline
        Machine Learning Model & 0.75 & 0.68 & 4.1 \\ \hline
        AltGen Pipeline & 0.93 & 0.76 & 4.8 \\ \hline
    \end{tabular}
    \caption{Comparison of AltGen with baseline methods.}
    \label{tab:comparative-analysis}
\end{table}

The comparative analysis, presented in Table~\ref{tab:comparative-analysis}, demonstrates the superiority of the AltGen pipeline over baseline methods for alt text generation. The \textbf{Rule-Based Approach}, while straightforward, achieved a Cosine Similarity of 0.65 and a BLEU Score of 0.55, reflecting its limited ability to produce semantically accurate and linguistically coherent alt text. User satisfaction with this method was also relatively low, averaging \textbf{3.2 out of 5}, primarily due to generic and contextually inadequate descriptions.

The \textbf{Machine Learning Model} improved upon the rule-based method, achieving a Cosine Similarity of 0.75 and a BLEU Score of 0.68. This approach leveraged pre-trained models to generate more accurate descriptions, leading to a higher user satisfaction rating of \textbf{4.1 out of 5}. However, its reliance on static features limited its adaptability to diverse content types within EPUB files.
In contrast, the \textbf{AltGen Pipeline} outperformed both baseline methods across all metrics. With a Cosine Similarity of 0.93 and a BLEU Score of 0.76, the pipeline demonstrated its ability to produce highly accurate and contextually relevant alt text. The user satisfaction score of \textbf{4.8 out of 5} underscores the practical value of AltGen in enhancing the usability of EPUB files for visually impaired users. Participants highlighted the pipeline’s capacity to generate descriptive and context-sensitive alt text that significantly improved their reading experience.

\section{Summary}

The AltGen pipeline demonstrates a novel approach to automating alt text generation for EPUB files, addressing critical accessibility gaps with remarkable efficiency and accuracy. By leveraging advanced generative AI models and incorporating robust evaluation metrics, the pipeline achieves high-quality, contextually relevant descriptions, evidenced by a Cosine Similarity of 0.93 and a BLEU Score of 0.76. User feedback underscores its impact, with visually impaired participants reporting significant improvements in usability and comprehension, achieving satisfaction ratings of 4.8 out of 5 for descriptiveness and 4.7 for overall usability. These results position AltGen as a transformative solution for digital inclusivity, setting a benchmark for future innovations in automated accessibility technologies.
\bibliography{acm}

\begin{thebibliography}{10}

\bibitem{achiam2023gpt}
{\sc Achiam, J., Adler, S., Agarwal, S., Ahmad, L., Akkaya, I., Aleman, F.~L., Almeida, D., Altenschmidt, J., Altman, S., Anadkat, S., et~al.}
\newblock Gpt-4 technical report.
\newblock {\em arXiv preprint arXiv:2303.08774\/} (2023).

\bibitem{aghapour2024piqi}
{\sc Aghapour, E., Shen, Y., Sapra, D., Pimentel, A., and Pathania, A.}
\newblock Piqi: Partially quantized dnn inference on hmpsocs.
\newblock In {\em Proceedings of the 29th ACM/IEEE International Symposium on Low Power Electronics and Design\/} (2024), pp.~1--6.

\bibitem{caldwell2008web}
{\sc Caldwell, B., Cooper, M., Reid, L.~G., Vanderheiden, G., Chisholm, W., Slatin, J., and White, J.}
\newblock Web content accessibility guidelines (wcag) 2.0.
\newblock {\em WWW Consortium (W3C) 290\/} (2008), 1--34.

\bibitem{celikyilmaz2020evaluation}
{\sc Celikyilmaz, A., Clark, E., and Gao, J.}
\newblock Evaluation of text generation: A survey.
\newblock {\em arXiv preprint arXiv:2006.14799\/} (2020).

\bibitem{chee2021meeting}
{\sc Chee, M., and Weaver, K.~D.}
\newblock Meeting a higher standard: A case study of accessibility compliance in libguides upon the adoption of wcag 2.0 guidelines.
\newblock {\em Journal of Web Librarianship 15}, 2 (2021), 69--89.

\bibitem{dosovitskiy2020image}
{\sc Dosovitskiy, A.}
\newblock An image is worth 16x16 words: Transformers for image recognition at scale.
\newblock {\em arXiv preprint arXiv:2010.11929\/} (2020).

\bibitem{edwards2023alt}
{\sc Edwards, E.~J., Gilbert, M., Blank, E., and Branham, S.~M.}
\newblock How the alt text gets made: What roles and processes of alt text creation can teach us about inclusive imagery.
\newblock {\em ACM Transactions on Accessible Computing 16}, 2 (2023), 1--28.

\bibitem{eikebrokk2014epub}
{\sc Eikebrokk, T., Dahl, T.~A., and Kessel, S.}
\newblock Epub as publication format in open access journals: tools and workflow.

\bibitem{elkhatat2023evaluating}
{\sc Elkhatat, A.~M., Elsaid, K., and Almeer, S.}
\newblock Evaluating the efficacy of ai content detection tools in differentiating between human and ai-generated text.
\newblock {\em International Journal for Educational Integrity 19}, 1 (2023), 17.

\bibitem{guo2024easter}
{\sc Guo, X., Jiang, Q., Shen, Y., Pimentel, A.~D., and Stefanov, T.}
\newblock Easter: Learning to split transformers at the edge robustly.
\newblock {\em IEEE Transactions on Computer-Aided Design of Integrated Circuits and Systems 43}, 11 (2024), 3626--3637.

\bibitem{hanley2021computer}
{\sc Hanley, M., Barocas, S., Levy, K., Azenkot, S., and Nissenbaum, H.}
\newblock Computer vision and conflicting values: Describing people with automated alt text.
\newblock In {\em Proceedings of the 2021 AAAI/ACM Conference on AI, Ethics, and Society\/} (2021), pp.~543--554.

\bibitem{hossain2019comprehensive}
{\sc Hossain, M.~Z., Sohel, F., Shiratuddin, M.~F., and Laga, H.}
\newblock A comprehensive survey of deep learning for image captioning.
\newblock {\em ACM Computing Surveys (CsUR) 51}, 6 (2019), 1--36.

\bibitem{huang2024optimizing}
{\sc Huang, J.-H., Yang, C.-C., Shen, Y., Pacces, A.~M., and Kanoulas, E.}
\newblock Optimizing numerical estimation and operational efficiency in the legal domain through large language models.
\newblock In {\em Proceedings of the 33rd ACM International Conference on Information and Knowledge Management\/} (2024), pp.~4554--4562.

\bibitem{huang2024image2text2image}
{\sc Huang, J.-H., Zhu, H., Shen, Y., Rudinac, S., and Kanoulas, E.}
\newblock Image2text2image: A novel framework for label-free evaluation of image-to-text generation with text-to-image diffusion models.
\newblock {\em arXiv preprint arXiv:2411.05706\/} (2024).

\bibitem{huang2024novel}
{\sc Huang, J.-H., Zhu, H., Shen, Y., Rudinac, S., Pacces, A.~M., and Kanoulas, E.}
\newblock A novel evaluation framework for image2text generation.
\newblock {\em arXiv preprint arXiv:2408.01723\/} (2024).

\bibitem{kulkarni2019digital}
{\sc Kulkarni, M.}
\newblock Digital accessibility: Challenges and opportunities.
\newblock {\em IIMB Management Review 31}, 1 (2019), 91--98.

\bibitem{laurenccon2024obelics}
{\sc Lauren{\c{c}}on, H., Saulnier, L., Tronchon, L., Bekman, S., Singh, A., Lozhkov, A., Wang, T., Karamcheti, S., Rush, A., Kiela, D., et~al.}
\newblock Obelics: An open web-scale filtered dataset of interleaved image-text documents.
\newblock {\em Advances in Neural Information Processing Systems 36\/} (2024).

\bibitem{liu2024ntire}
{\sc Liu, X., Min, X., Zhai, G., Li, C., Kou, T., Sun, W., Wu, H., Gao, Y., Cao, Y., Zhang, Z., et~al.}
\newblock Ntire 2024 quality assessment of ai-generated content challenge.
\newblock In {\em Proceedings of the IEEE/CVF Conference on Computer Vision and Pattern Recognition\/} (2024), pp.~6337--6362.

\bibitem{marinai2011conversion}
{\sc Marinai, S., Marino, E., and Soda, G.}
\newblock Conversion of pdf books in epub format.
\newblock In {\em 2011 International Conference on Document Analysis and Recognition\/} (2011), IEEE, pp.~478--482.

\bibitem{paul2023accessibility}
{\sc Paul, S.}
\newblock Accessibility analysis using wcag 2.1: evidence from indian e-government websites.
\newblock {\em Universal access in the information society 22}, 2 (2023), 663--669.

\bibitem{radford2021learning}
{\sc Radford, A., Kim, J.~W., Hallacy, C., Ramesh, A., Goh, G., Agarwal, S., Sastry, G., Askell, A., Mishkin, P., Clark, J., et~al.}
\newblock Learning transferable visual models from natural language supervision.
\newblock In {\em International conference on machine learning\/} (2021), PMLR, pp.~8748--8763.

\bibitem{shen2024parameter}
{\sc Shen, Y., Bi, Q., Huang, J.-H., Zhu, H., and Pathania, A.}
\newblock Parameter-efficient fine-tuning via selective discrete cosine transform.
\newblock {\em arXiv preprint arXiv:2410.09103\/} (2024).

\bibitem{shen2024resource}
{\sc Shen, Y., et~al.}
\newblock {\em Resource-aware scheduling for 2D/3D multi-/many-core processor-memory systems}.
\newblock Yixian Shen, 2024.

\bibitem{shen2022tcps}
{\sc Shen, Y., Xiao, J., and Pimentel, A.~D.}
\newblock Tcps: a task and cache-aware partitioned scheduler for hard real-time multi-core systems.
\newblock In {\em Proceedings of the 23rd ACM SIGPLAN/SIGBED International Conference on Languages, Compilers, and Tools for Embedded Systems\/} (2022), pp.~37--49.

\bibitem{vaswani2017attention}
{\sc Vaswani, A.}
\newblock Attention is all you need.
\newblock {\em Advances in Neural Information Processing Systems\/} (2017).

\bibitem{vaughan2010will}
{\sc Vaughan-Nichols, S.~J.}
\newblock Will html 5 restandardize the web?
\newblock {\em Computer 43}, 4 (2010), 13--15.

\bibitem{wu2017automatic}
{\sc Wu, S., Wieland, J., Farivar, O., and Schiller, J.}
\newblock Automatic alt-text: Computer-generated image descriptions for blind users on a social network service.
\newblock In {\em proceedings of the 2017 ACM conference on computer supported cooperative work and social computing\/} (2017), pp.~1180--1192.

\end{thebibliography}
\bibliographystyle{acm}
\end{document}